\documentclass{article}

\usepackage[preprint]{neurips_2026}

\usepackage[utf8]{inputenc} 
\usepackage[T1]{fontenc}    
\usepackage{hyperref}       
\usepackage{url}            
\usepackage{booktabs}       
\usepackage{amsfonts}       
\usepackage{nicefrac}       
\usepackage{microtype}      
\usepackage[table]{xcolor}         
\usepackage{colortbl}
\definecolor{tablelightgray}{gray}{0.9}

\usepackage{amsmath}
\usepackage{algorithm}
\usepackage{algpseudocode}
\usepackage{enumitem}
\usepackage{cleveref}

\usepackage{multirow}
\usepackage{array}
\usepackage{graphicx}
\usepackage{booktabs}

\title{Closed Loop Dynamic Driving Data Mixture \\ for Real-Synthetic Co-Training}

\author{
    \textbf{Hongzhi Ruan} $^{1,}\footnotemark[2]$ \qquad
    \textbf{Pei Liu} $^{1,3}$ \qquad
    \textbf{Weiliang Ma} $^{1}$ \qquad
    \textbf{Zhengning Li} $^{1}$ \\[1mm]
    \textbf{Xueyang Zhang} $^{1}$ \qquad
    \textbf{Jun Ma} $^{3}$ \qquad
    \textbf{Dan Xu} $^{2}$ \qquad
    \textbf{Kun Zhan} $^{1}$ \\[3mm]
    $^{1}$ Li Auto \qquad
    $^{2}$ HKUST \qquad
    $^{3}$ HKUST (GZ)
}

\begin{document}

\maketitle

\vspace{-5mm}
\renewcommand{\thefootnote}{\fnsymbol{footnote}}
\footnotetext[2]{\texttt{ruanhongzhicn@gmail.com}}

\begin{abstract}

Data scaling is fundamental to modern deep learning, and grows increasingly critical as autonomous driving shifts to end-to-end learning.
Real-world driving data is expensive to annotate and scene-biased, making real-synthetic co-training with near-infinite synthetic data a promising direction.
However, naively incorporating all available synthetic data is inefficient and leads to distribution shifts, and optimizing data mixture under practical training budgets remains a critical yet under-explored problem. In this sense, we claim that the mixture of training data requires clear guidance in terms of scene types and quantities.
Particularly in this work, we conceptualize the data mixture approximately as a dynamic optimization process that iteratively adjusts the training data mixture to maximize model performance, guided by closed-loop evaluation feedback, and propose AutoScale, a fully automated closed-loop data engine unifying scene representation, data mixture optimization and retrieval, as well as model training and evaluation.
Specifically, we propose Graph Regularized AutoEncoder (Graph-RAE) for driving scene representations, introduce Cluster-aware Gradient Ascent (Cluster-GA) for cluster-wise importance estimation and reweighting, and perform cluster-guided vector retrieval to select high-value samples.
Experiments on NavSim demonstrate that AutoScale outperforms vanilla co-training and cross-domain baselines, achieving better performance with fewer synthetic samples under constrained budgets.

\end{abstract}

\section{Introduction}

Data scaling has become a fundamental principle for building large modern deep learning models across language, vision, and generative tasks~\cite{brown2020gpt3,kaplan2020llmscaling,dosovitskiy2020vit,radford2021clip,blattmann2023svd}.
As autonomous driving evolves from modular pipelines toward end-to-end learning from raw sensor data to actions~\cite{hu2023uniad,chen2024e2e}, data scaling laws have grown increasingly critical in this domain.

Real-world driving data, while reliable, is inherently limited, expensive to annotate, and biased toward frequent, safe driving scenarios.
To improve data diversity and coverage, recent research has begun to focus on synthetic data, which can be generated in large quantities at low cost.
Leveraging generation or reconstruction techniques, recent works attempt to generate a large amount of synthetic data based on prior scenes, and further achieve performance improvement through real-synthetic co-training~\cite{ren2025cosmos,ge2025unravelscale,tian2025simscale}.
However, existing works mostly focus on generating more and better synthetic data, while few have systematically studied how to optimally mix them with real data.

\begin{figure}
    \centering
    \includegraphics[width=1.0\linewidth]{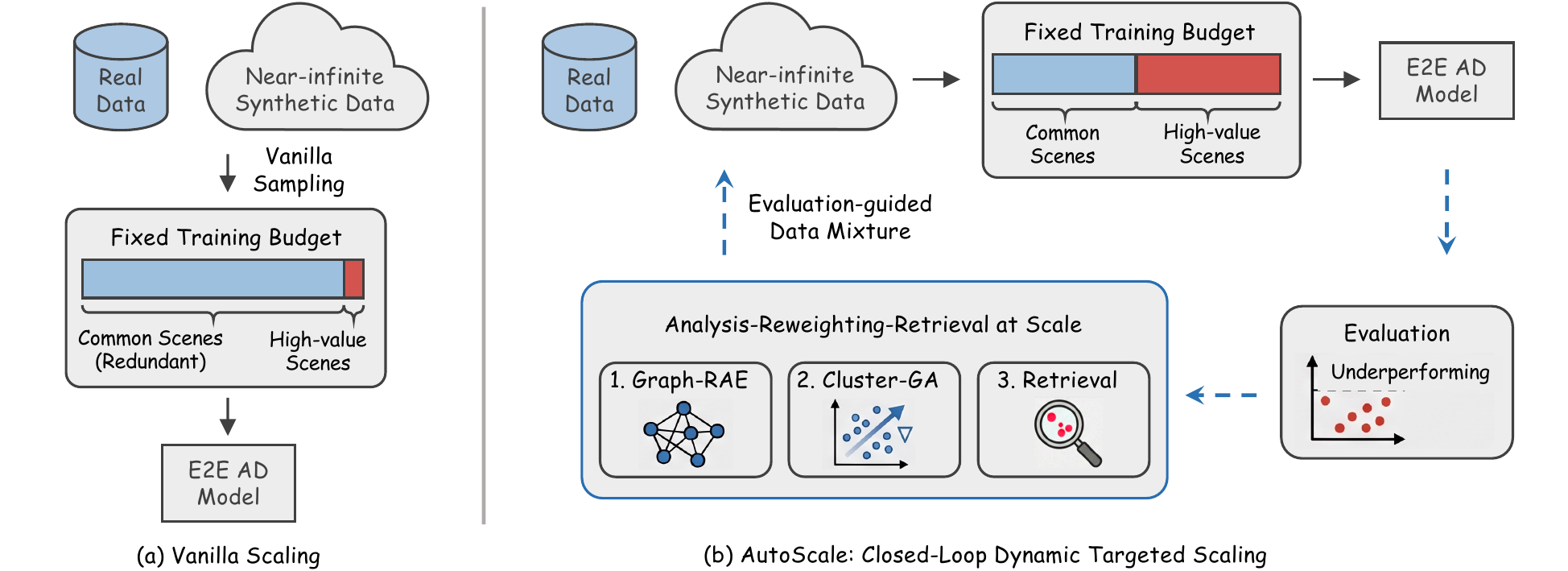}
    \vspace{-3mm}
    \caption{Scheme Comparison.
    (a) Existing scaling for real-synthetic co-training adopts vanilla sampling from nearly infinite unguided synthetic data.
    (b) Our AutoScale closed-loop framework leverages evaluation feedback to enable scalable scene analysis, cluster-wise reweighting, and targeted synthetic data retrieval, providing explicit guidance for effective data mixture optimization.}
    \vspace{-3mm}
    \label{fig:scheme_comparison}
\end{figure}

Under practical training budgets, it is critical to optimize the data mixture and precisely identify valuable samples for model improvement, as the amount of data the model can effectively learn from is inherently limited~\cite{baniodeh2025waymoscale,zheng2024datascaling}.
In practice, naively injecting all available synthetic data is inefficient and often degrades performance due to redundancy, noise, and distribution mismatch~Fig.~\ref{fig:scheme_comparison} (a).
This problem remains under-explored because of several key challenges:
(1) Diverse driving scenes are structurally complex and difficult to analyze and localize at scale;
(2) It is difficult to quantify the impact of the types and proportions of different driving scenes in the training set on model performance, which hinders the optimization of the training data mixture;
(3) It is challenging to use model failure cases from evaluation to targetedly enhance the training distribution.

In this paper, we present a pioneering systematic investigation of the data mixture effect on overall model performance for real-synthetic co-training.
We adopt a practical and realistic setting: a fixed, well-distributed real dataset is preserved without modification, and a large-scale pre-generated external dataset serves as an accessible, low-cost candidate pool, from which we sample a fixed-budget set of synthetic data to construct a superior mixture for real-synthetic co-training.
We argue that the mixture of training data requires clear guidance in terms of scene types and quantities.
With this setting, we further formulate the incorporation of synthetic data approximately as a gradient ascent optimization process toward improved model performance.
In practice, we introduce AutoScale~Fig.~\ref{fig:scheme_comparison} (b), a closed-loop dynamic data engine that unifies scene representation, data mixture optimization and retrieval, model training, and evaluation for real-synthetic co-training.

Specifically, to precisely address the aforementioned challenges, we first propose Graph Regularized AutoEncoder (Graph-RAE), which encodes and reconstructs driving scene information and integrates contrastive and metric learning regularization to represent driving scenes as vectors.
Second, we cluster driving scenes using these learned scene vectors, and further introduce Cluster-aware Gradient Ascent (Cluster-GA) to iteratively estimate cluster-wise importance weights within the training-evaluation closed loop for data reweighting.
Finally, guided by the learned cluster importance, we leverage scene vectors to retrieve high-value synthetic samples that compensate for model weaknesses.

We conduct extensive experiments on the NavSim~\cite{Dauner2024navsim} benchmark.
We use the training subset from the navtrain split as our fixed real dataset, and adopt the large-scale SimScale~\cite{tian2025simscale} dataset generated via 3D Gaussian Splatting~\cite{kerbl20233dgs,li2025mtgs}, from which we split a held-out calibration set for closed-loop evaluation and use the remainder as our synthetic candidate pool.
Evaluated on the regression-based planner LTF~\cite{Chitta2023transfuser} and diffusion-based planner DiffusionDrive~\cite{liao2025diffusiondrive}, AutoScale utilizes simulation evaluation guidance to outperform vanilla co-training and cross-domain methods, achieving better real-world benchmark results with fewer synthetic samples.
Ablation studies and comprehensive analyses verify the effectiveness of Graph-RAE, Cluster-GA, and vector-based retrieval.

Our contributions are fourfold:
\begin{itemize}[leftmargin=*, itemindent=0pt]
    \item We pioneer the optimization of data mixtures for improved model performance in real-synthetic co-training, and propose a fully automated closed-loop data engine.
    \item We propose Graph-RAE, a Graph Regularized AutoEncoder integrated with contrastive and metric learning, enabling effective representation and efficient analysis of driving scenes at scale.
    \item We introduce Cluster-GA, a gradient ascent-based method for driving scene cluster reweighting.
    \item We perform vector-based scene retrieval to retrieve high-value synthetic samples for training.
\end{itemize} 
\section{Related Work}

\textbf{Data Centric Autonomous Driving.}
With the shift of autonomous driving from modular pipelines to end-to-end learning that maps raw sensor data to driving actions~\cite{hu2023uniad,chen2024e2e}, data distribution learning has become increasingly critical. 
Representative methods include regression-based planners~\cite{Chitta2023transfuser,jiang2023vad,sun2025sparsedrive} and generation-based planners~\cite{liao2025diffusiondrive,xing2025goalflow,li2025recogdrive,zheng2025diffusionplanner}, both of which rely heavily on high-quality training data. 
On the other hand, since real-world driving data is expensive to annotate and scene-biased, recent research has turned its attention to synthetic data with inherently precise annotation and near-infinite generation at low cost.
Existing synthetic data methods can be divided into two categories: generation-based approaches~\cite{hu2023gaia,gao2024vista,ma2024delphi,ding2024realgen,wen2024panacea,gao2023magicdrive,wang2024drivedreamer,gao2025magicdrivev2} and reconstruction-based approaches~\cite{lin2025mpadrive,garcia2025road,tian2025simscale}. 
Beyond using synthetic data alone, real-synthetic co-training has been proven to achieve better model performance than single-data-source training~\cite{tian2025simscale,ren2025cosmos,ge2025unravelscale}. 
However, most existing works focus on generating more and higher-quality synthetic data; despite the availability of near-infinite synthetic data, few studies have systematically investigated how to optimally mix synthetic data with real data under practical training budgets, this gap is the core focus of our work.

\textbf{Data Mixtures.}
The composition of pre-training datasets is critical in determining the generalization ability of large language models (LLMs)~\cite{brown2020gpt3,touvron2023llama}.
As a result, a large body of research has focused on data mixture optimization in LLMs~\cite{xie2023doremi,fan2024doge,liu2025regmix,diao2025climb,ye2025datamixinglaws}. Similarly, the mixture of real and synthetic data is also a critical problem in action imitation learning for robotics~\cite{gr00tn1_2025,maddukuri2025simandreal,Nasiriany2024robocasa,team2025gigaworld,tian2025interndata}.
Despite its importance, data mixture optimization remains under-explored in both robotics and autonomous driving~\cite{baniodeh2025waymoscale,zheng2024datascaling}, primarily due to the structural complexity of representing both states and actions, which hinders large-scale analysis and localization of key data patterns.
To highlight the advantages of our proposed AutoScale, we select two representative data mixture methods as baselines.
Chameleon~\cite{xie2025chameleon}, a data mixture framework designed for LLMs, leverages domain text embeddings to assign mixture weights to each domain, achieving superior performance on language benchmarks.
IWR~\cite{xie2025iwr} introduces Importance Weighted Retrieval, which estimates the importance weights of target data samples within data distributions and measures robotics task similarity in the latent space to guide sample retrieval.
In this work, we first present Graph-RAE to encode driving scenes into structured vectors, addressing the challenge of complex driving scene representation.
Further, unlike most existing data mixture baselines that rely solely on training data distribution analysis, we optimize data mixture through cluster-wise reweighting and targeted sample retrieval, guided by closed-loop model performance evaluation feedback.

\section{Task Definition}
\label{sec:task}

We consider an end-to-end imitation learning setting for autonomous driving, where the model is trained on a mixed dataset consisting of fixed real-world data and selectively augmented synthetic driving scenes.
Let $\mathcal{D}_{\mathrm{real}}$ denote the real training set with $N_0 = |\mathcal{D}_{\mathrm{real}}|$ samples, and $\mathcal{D}_{\mathrm{syn}}$ represent a large candidate pool of synthetic data generated by specialized synthesis pipelines.
Given a fixed synthetic data budget $B$, the total size of the final mixed training set is $N = N_0 + B$.

Each driving scene is treated as a data sample $x$, and its contextual information is encoded into a structured vector $\mathbf{v}_x \in \mathbb{R}^d$ via our proposed Graph-RAE (Section~\ref{sec:graph_rae}).
A held-out calibration split $\mathcal{D}_{\mathrm{cal}}$ is adopted to evaluate the driving policy $\pi_\theta$ with closed-loop driving metrics, producing a comprehensive performance score $S$.
In our setting, all real data is fully preserved, while only synthetic samples are selected to fill the predefined budget.

\textbf{Target.}
Our objective is to select optimal $B$ synthetic samples from $\mathcal{D}_{\mathrm{syn}}$ to augment $\mathcal{D}_{\mathrm{real}}$ and refine the real-synthetic data mixture.
Guided by simulation evaluation score $S$ as quantitative feedback, we aim to enhance overall model performance on the real-world test set.
\section{Methodology}
\label{sec:method}

\begin{figure}
    \centering
    \includegraphics[width=1.0\linewidth]{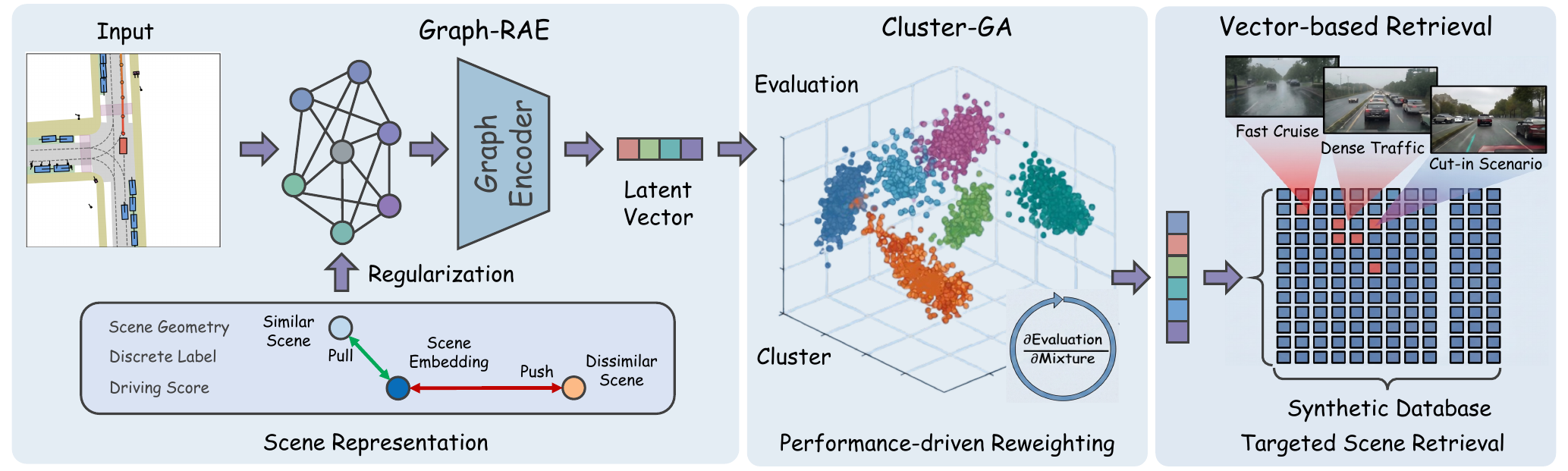}
    \vspace{-6mm}
    \caption{Overview of the proposed AutoScale with three core modules. 
    (a) Scene representation learning: Graph-RAE encodes complex driving scenes into structured vectors. 
    (b) Cluster-level weight optimization: Cluster-GA estimates cluster-wise data importance based on closed-loop evaluation feedback. 
    (c) Targeted synthetic data retrieval: Vector-based retrieval selects high-value synthetic samples to adjust the training distribution for model underperformance.}
    \label{fig:autoscale}
\end{figure}

This section presents AutoScale, a closed-loop dynamic data engine that unifies scene representation, data mixture optimization, retrieval, model training, and evaluation for real-synthetic co-training.
The three core modules in the framework are depicted in \cref{fig:autoscale}.


\subsection{Graph Regularized Autoencoder (Graph-RAE)}
\label{sec:graph_rae}

We learn a scene-level representation $\mathbf{v}_x \in \mathbb{R}^d$
that jointly captures the static map structure, dynamic traffic participants, and driving behavior of the ego vehicle within each traffic scene.

\textbf{Graph Spatial-Temporal Autoencoder.}
We model each driving scenario as a heterogeneous spatio-temporal graph constructed over $T_{\mathrm{len}}$ sequential frames ($T_h$ historical + $T_f$ future).
The graph consists of three distinct node categories: the ego-vehicle, dynamic traffic participants (vehicles, pedestrians, cyclists, etc.), and map structural elements (ped\_crossing, divider, boundary).
Each node is associated with a $T_{\mathrm{len}}$-step polyline sequence comprising 2D position, heading, and curvature.
Directed edge types are designed to encode pairwise spatial interactions, including map-to-dynamic, dynamic-to-dynamic, and all-to-ego relational dependencies; each edge encodes the relative position, heading difference, and curvature deviation between connected nodes across all $T_{\mathrm{len}}$ steps.

Node features at each step are encoded via shared multi-layer perceptrons (MLPs), augmented with learnable positional embeddings and node-type embeddings, followed by sequential aggregation to produce a node representation $\mathbf{h}_i \in \mathbb{R}^d$.
Edge features are encoded through an identical pipeline.
An $L$-layer Graph Transformer then refines all node representations via gated multi-head attention~\cite{vaswani2017attention}, formulated as:
\begin{equation}
  \mathbf{h}_i^{(l+1)}
  = \mathbf{h}_i^{(l)} + \sigma\!\big(\mathbf{g}_i^{(l)}\big)
    \odot \mathrm{MHA}\!\big(
      \mathbf{h}_i^{(l)},\;
      \{\mathbf{h}_j^{(l)},\, \mathbf{e}_{ji}\}_{j \in \mathcal{N}(i)}
    \big),
  \label{eq:gnn}
\end{equation}
where $\sigma\big(\mathbf{g}_i^{(l)}\big)$ denotes a learnable gating function and $\mathrm{MHA}$ computes multi-head attention over neighboring node features concatenated with edge attributes.
The global scene representation is derived from the $\ell_2$-normalized embedding of the ego-vehicle node after the final layer and enhanced with regularization.
Finally, a graph decoder reconstructs the complete polyline sequences of all nodes from the learned representations, optimized using the Huber loss.

\textbf{Contrastive Learning for Regularization.}\quad
To regularize the embedding space with traffic semantics, we employ two contrastive regularization objectives.
First, the SimCLR~\cite{chen2020simclr} objective enforces invariance to geometric perturbations:
two augmented views of the same scene, generated by random translation and rotation of all polylines, are treated as positive pairs, while all other scenes serve as negative samples.
Second, the supervised contrastive (SupCon)~\cite{khosla2020supcon} objective clusters scenes according to discrete semantic labels, pulling embeddings of scenes with consistent human-annotated priors closer together.
Here, we simply define two semantic annotation types:
a command pattern label that characterizes the ego-vehicle’s high-level driving intent, and a geometric overlap label that indicates whether the ego trajectory intersects with dynamic agents or map elements.
This design allows seamless integration of additional semantic labels for enhanced representation learning.

\textbf{Metric Learning for Regularization.}\quad
To further align the embedding space with driving task semantics, we introduce a pairwise metric regression objective.
For each pair of scenes $(x_i, x_j)$ in a batch, their ego embeddings are projected via an MLP and compared. 
We run training samples with evaluation metrics, and adopt a regression head to predict the element-wise absolute difference of evaluation metric vectors $|\mathbf{m}_i - \mathbf{m}_j| \in \mathbb{R}^{N_m}$, where $N_m$ denotes the dimension of evaluation metrics.
The objective is supervised by the mean squared error (MSE) loss over all valid pairs.
This objective ensures that scenes with analogous driving performance scores are embedded in proximity, directly benefiting downstream driving analysis tasks.

The overall training loss is formulated as:
\begin{equation}
\mathcal{L}
= \mathcal{L}_{\mathrm{recon}}
+ \lambda \left(
\mathcal{L}_{\mathrm{contra}}
+ \mathcal{L}_{\mathrm{metric}}
\right).
\label{eq:total_loss}
\end{equation}

\subsection{Gradient Ascent Cluster Reweighting}
We first partition driving scenes into distinct clusters, and formulate the relationship between cluster-wise data mixture and evaluation performance as a gradient ascent optimization paradigm (Cluster-GA). A predictor is further introduced to iteratively model this optimization process.

\textbf{Data Clustering.}\quad
Using the scene embeddings from Graph-RAE, we partition all real training tokens into $K$ clusters
$\{C_1, \dots, C_K\}$ via Gaussian Mixture Models (GMM) on $\ell_2$-normalized vectors, yielding cluster centroids $\{\mathbf{c}_1, \dots, \mathbf{c}_K\}$.
Let $n_{0,k}$ denote the number of real tokens in cluster~$k$, and define the data mixture vector $\mathbf{w} \in \Delta_K$, where $w_k$ is the fraction of total samples $N$ allocated to cluster~$k$.
The calibration set is assigned to the same clustering partition with a fixed mixture $\boldsymbol{\pi} \in \Delta_K$, enabling per-cluster evaluation feedback.
The evaluation metric produces per-cluster scores $\bar{s}_k$, and the overall evaluation performance is equivalently formulated as $S(\mathbf{w}) = \boldsymbol{\pi}^\top \bar{\mathbf{s}}(\mathbf{w})$.
To capture inter-cluster transfer effects, since data from one cluster can benefit semantically similar clusters, we adopt a precomputed fixed similarity matrix $\mathbf{R} \in \mathbb{R}^{K\times K}$, whose entries follow the RBF kernel: $R_{kj}=\exp\big(-(1-\cos(\mathbf{c}_k,\mathbf{c}_j))^2/(2\sigma^2)\big)$.

\textbf{Cluster Performance Predictor.}\quad
Since the evaluation score $S(\mathbf{w})$ requires a full train–evaluate closed-loop to measure, we fit a model that predicts how changes in data mixture affect per-cluster scores.
Given two observed rounds $(a, b)$ with mixtures $\mathbf{w}^{(a)}, \mathbf{w}^{(b)}$ and scores $\bar{\mathbf{s}}^{(a)}, \bar{\mathbf{s}}^{(b)}$, the per-cluster score difference is modeled as:
\begin{equation}
  \Delta\bar{s}_k^{(a \to b)}
  = \sum_{j=1}^{K} R_{kj}\,\beta_j\,
    \Delta \log w_j
  + \gamma \sum_{j=1}^{K} R_{kj}\,\Delta r_j,
  \label{eq:surrogate}
\end{equation}
where $\Delta \log w_j = \log w_j^{(b)} - \log w_j^{(a)}$ and $\Delta r_j = r_j(\mathbf{w}^{(b)}) - r_j(\mathbf{w}^{(a)})$.
$r_j(\mathbf{w}) = \max(0,\, 1 - n_{0,j}/(w_j N))$ denotes the synthetic data ratio for cluster $j$.
$\beta_j$ captures the marginal data return in cluster~$j$ (the logarithmic term reflects diminishing returns), and $\gamma$ scales synthetic data effects.

In each iteration, given $t$ historical rounds, we construct all $\binom{t}{2}$ pairwise samples and solve for parameters $\boldsymbol{\phi} = [\boldsymbol{\beta}^\top;\, \gamma] \in \mathbb{R}^{K+1}$ via weighted ridge regression in closed form:
\begin{equation}
  \boldsymbol{\phi}
  = \big(\mathbf{X}^\top \mathbf{W} \mathbf{X}
         + \lambda_{\mathrm{reg}} \mathbf{I}\big)^{-1}
    \mathbf{X}^\top \mathbf{W}\, \mathbf{y}.
  \label{eq:fit}
\end{equation}
where $\mathbf{X}$ and $\mathbf{y}$ are the feature matrix and regression target vector, $\mathbf{W}$ is a diagonal weight matrix, $\mathbf{I}$ denotes the identity matrix, and $\lambda_{\mathrm{reg}}$ controls regularization strength.

\textbf{Gradient Ascent Optimization.}\quad
At round~1, only the baseline observation $(\mathbf{w}^{(0)}, \bar{\mathbf{s}}^{(0)})$ is available, which is insufficient to fit the performance predictor.
We set the cluster gain vector $\boldsymbol{\alpha} = \mathbf{0}$ and rely on pure token-level retrieval (Section~\ref{sec:retrieval}) to select $B$ synthetic tokens.
The resulting data composition naturally induces a new mixture $\mathbf{w}^{(1)}$ that reflects the policy’s failure modes: clusters with poor model performance receive more synthetic data via the score-weighted priority in~\eqref{eq:priority}.
This observation pair $(\mathbf{w}^{(1)}, \bar{\mathbf{s}}^{(1)})$, together with the baseline, is used to train the predictor at round~2.

Starting from round~2, we adopt exponentiated gradient (EG) ascent to maximize the predictor performance $\hat{S}(\mathbf{w})$.
Each optimization step is initialized from the weight vector $\mathbf{w}^{(t^*)}$ with the best historical evaluation score, where $t^* = \arg\max_{\ell < t}\boldsymbol{\pi}^\top \bar{\mathbf{s}}^{(\ell)}$.
This design prevents performance drift caused by predictor inaccuracy.
The gradient of $\hat{S}$ with respect to $\theta_j = \log w_j$ is formulated as:
\begin{equation}
  g_j
  = (\boldsymbol{\pi}^\top \mathbf{R})_j
    \left(
      \beta_j
      + \gamma \,\frac{n_{0,j}}{w_j\,N}
    \right),
  \label{eq:gradient}
\end{equation}
and the cluster mixture is updated in a multiplicative EG manner:
\begin{equation}
  w_k^{(t)}
  \leftarrow
  \frac{w_k^{(\mathrm{prev})} \exp(\eta_t\, g_k)}
       {\sum_j w_j^{(\mathrm{prev})} \exp(\eta_t\, g_j)},
  \label{eq:eg_update}
\end{equation}
where the step size $\eta_t$ is calibrated to ensure $\|\mathbf{w}^{(t)} - \mathbf{w}^{(\mathrm{prev})}\|_2$ follows a half-cosine schedule $\epsilon_t$.
If the updated mixture violates global data constraints, a lightweight perturbed search over the feasible set is adopted to obtain a valid candidate.
The per-cluster augmentation size is finally determined as $\delta_k = \max(\lfloor w_k^{(t)} N \rceil - n_{0,k},\, 0)$.

Meanwhile, the predictor yields a per-cluster performance gain $\boldsymbol{\alpha}$ in score space.
It measures the predicted performance improvement from the previous mixture to the updated one, and is normalized into $[0, 1]$ via:
\begin{equation}
  \alpha_k
  = \frac{
      \max\!\big(
        \hat{\bar{s}}_k(\mathbf{w}^{(t)})
        - \hat{\bar{s}}_k(\mathbf{w}^{(\mathrm{prev})}),
        \;0\big)
    }{
      \max_j\!\big(
        \hat{\bar{s}}_j(\mathbf{w}^{(t)})
        - \hat{\bar{s}}_j(\mathbf{w}^{(\mathrm{prev})})
      \big)
    }.
  \label{eq:gain}
\end{equation}
This gain $\boldsymbol{\alpha}$ is fed into the downstream data retrieval stage as a weighting factor (Section~\ref{sec:retrieval}).

\begin{algorithm}[t]
\caption{AutoScale Data Engine}
\label{alg:autoscale}
\begin{algorithmic}[1]
\Require
  Real data $\mathcal{D}_{\mathrm{real}}$ with cluster counts $\mathbf{n}_0$;
  synthetic pool $\mathcal{D}_{\mathrm{syn}}$;
  budget $B$; total rounds $T$;
  step schedule $\{\epsilon_t\}_{t=1}^{T}$

\Ensure Optimized policy $\pi_{\theta^*}$

\State Extract scene embeddings via Graph-RAE;\;
       cluster embeddings with GMM to obtain $K$ clusters $\{C_k\}$ and centroids $\{\mathbf{c}_k\}$
       \Comment{Clustering \& Representation}
\State $N \leftarrow N_0 + B$;\;
       $\mathbf{w}^{(0)} \leftarrow \mathbf{n}_0 / N_0$;\;
       compute $\boldsymbol{\pi}$ and $\mathbf{R}$;
       $\mathcal{W} \leftarrow \{\mathbf{w}^{(0)}\}$
       \Comment{Initialization}

\State Train $\theta_0$ on $\mathcal{D}_{\mathrm{real}}$;\;
       evaluate $\bar{\mathbf{s}}^{(0)}$
       \Comment{Round 0: Baseline}

\State $\boldsymbol{\alpha} \leftarrow \mathbf{0}$
       \Comment{Round 1: Heuristical retrieval}
\State $\mathcal{D}_{\mathrm{add}}^{(1)} \leftarrow$
       \Call{Retrieve}{$\boldsymbol{\alpha},\, B,\, \mathcal{D}_{\mathrm{syn}},\, \mathcal{D}_{\mathrm{cal}}$}
       \Comment{Data Retrieval}
\State Train $\theta_1$ on $\mathcal{D}_{\mathrm{real}} \cup \mathcal{D}_{\mathrm{add}}^{(1)}$;\;
       evaluate $\bar{\mathbf{s}}^{(1)}$;\;
       record $\mathbf{w}^{(1)}$;
       $\mathcal{W} \leftarrow \mathcal{W} \cup \{\mathbf{w}^{(1)}\}$

\For{$t = 2, \dots, T$}
  \Comment{Round 2+: Cluster-GA Optimization}
  \State Fit $(\boldsymbol{\beta}, \gamma)$ via Eq.~\eqref{eq:fit} on pairs in $\mathcal{W}$
         \Comment{Cluster Performance Predictor}
  \State $t^* \!\leftarrow \arg\max_{\ell < t}\; \boldsymbol{\pi}^\top \bar{\mathbf{s}}^{(\ell)}$;\;
         $\mathbf{w}^{(\mathrm{prev})} \!\leftarrow \mathbf{w}^{(t^*)}$
  \State Compute gradient $\mathbf{g}$ via Eq.~\eqref{eq:gradient};\;
         update mixture $\mathbf{w}^{(t)}$ via Eq.~\eqref{eq:eg_update}
         \Comment{Optimize weight $\mathbf{w}^{(t)}$}
  \State $\boldsymbol{\delta} \leftarrow \max(\lfloor \mathbf{w}^{(t)} N \rceil - \mathbf{n}_0,\, \mathbf{0})$;\;
         compute gain $\boldsymbol{\alpha}$ via Eq.~\eqref{eq:gain}
  \State $\mathcal{D}_{\mathrm{add}}^{(t)} \leftarrow$
         \Call{Retrieve}{$\boldsymbol{\alpha},\, B,\, \mathcal{D}_{\mathrm{syn}},\, \mathcal{D}_{\mathrm{cal}}$}
         \Comment{Data Retrieval}
  \State Train $\theta_t$ on $\mathcal{D}_{\mathrm{real}} \cup \mathcal{D}_{\mathrm{add}}^{(t)}$;\;
         evaluate $\bar{\mathbf{s}}^{(t)}$;
         $\mathcal{W} \leftarrow \mathcal{W} \cup \{\mathbf{w}^{(t)}\}$
\EndFor

\State \Return $\pi_{\theta^*}$
\end{algorithmic}
\end{algorithm}

\subsection{Data Retrieval}
\label{sec:retrieval}

Given the total budget $B$ and cluster gain $\boldsymbol{\alpha}$ from Cluster-GA, we retrieve concrete synthetic tokens from $\mathcal{D}_{\mathrm{syn}}$ via a matching strategy that combines cluster-level guidance with token-level similarity.

For each cluster $k$, we identify anchor tokens from calibration scenes in $C_k$ where the current policy underperforms.
For each anchor $x_i \in C_k$ and candidate $x_j \in \mathcal{D}_{\mathrm{syn}}$, a retrieval priority is computed:
\begin{equation}
  p(x_i, x_j)
  = \big(1 + \alpha_{c(x_i)}\big)
    \cdot
    \big(1 - \bar{s}(x_i)\big)
    \cdot
    \cos(\mathbf{v}_{x_i},\, \mathbf{v}_{x_j}),
  \label{eq:priority}
\end{equation}
where $c(x_i)$ denotes the cluster assignment of anchor $x_i$, and $\alpha_{c(x_i)} \in [0, 1]$ is the normalized gain from Equation~\eqref{eq:gain}.
The three factors serve complementary roles: $(1 + \alpha_{c(x_i)})$ provides a cluster-level boost from the predictor;
$(1 - \bar{s}(x_i))$ amplifies the demand from low-scoring anchors within each cluster;
and $\cos(\mathbf{v}_{x_i}, \mathbf{v}_{x_j})$ ensures representational proximity in the Graph-RAE embedding space.

All anchor--candidate pairs across all clusters are pooled into a single global ranking by $p$, and the top-$B$ synthetic tokens are selected.
The final training set for round $t$ is $\mathcal{D}^{(t)} = \mathcal{D}_{\mathrm{real}} \cup \mathcal{D}_{\mathrm{add}}^{(t)}$.
\section{Experiments}
\label{sec:experiments}

In this section, we evaluate the AutoScale data engine on real-synthetic co-training tasks, and compare it with vanilla co-training strategies and cross-domain data mixture methods.
We also conduct experiments and discussions to verify the effectiveness of key designs of our method.

\begin{table}[t!]
  \centering
  \caption{Comparison of Real-Synthetic Co-Training in the NAVSIM-v2 \texttt{navhard} Benchmark.}
  \label{tab:real_synthetic_cotraining_navhard}
  \resizebox{\linewidth}{!}{%
  \begin{tabular}{l|c|c|c|c|cccc|ccccc|>{\columncolor{tablelightgray}}c}
    \toprule[1.2pt]
    \textbf{Model} & \textbf{Settings} & \textbf{Real} & \textbf{Synthetic} 
    & \textbf{Stage} & \textbf{NC↑} & \textbf{DAC↑} & \textbf{DDC↑} & \textbf{TLC↑} 
    & \textbf{EP↑} & \textbf{TTC↑} & \textbf{LK↑} & \textbf{HC↑} & \textbf{EC↑} & \cellcolor{white}\textbf{EPDMS↑} \\
    \midrule
    \multirow{10}{*}{TransFuser~\cite{Chitta2023transfuser}}
    & \multirow{2}{*}{Real Data}          & \multirow{2}{*}{85K}  & \multirow{2}{*}{-}    & S1 & 97.3 & 80.2 & 97.8 & 99.3 & 83.4 & 96.2 & 92.9 & 97.8 & 71.1 &  \\   &   &   &
    & S2 & 79.4 & 69.0 & 85.6 & 98.5 & 83.8 & 76.7 & 47.9 & 97.0 & 70.6 & \multirow{-2}{*}{24.4} \\
    \cmidrule{2-15}
    & \multirow{2}{*}{SimScale-recovery} & \multirow{2}{*}{85K}  & \multirow{2}{*}{147K} & S1 & 96.4 & 78.4 & 98.9 & 99.8 & 80.5 & 96.2 & 92.7 & 97.6 & 78.2 &  \\
    &                                    &                       &                       & S2 & 88.9 & 71.1 & 91.8 & 99.0 & 77.3 & 85.5 & 53.8 & 96.8 & 47.5 & \multirow{-2}{*}{29.8} \\
    \cmidrule{2-15}
    & \multirow{2}{*}{SimScale-planner}  & \multirow{2}{*}{85K}  & \multirow{2}{*}{237K} & S1 & 96.1 & 85.3 & 99.4 & 99.3 & 84.7 & 94.7 & 93.6 & 97.6 & 77.3 &  \\
    &                                    &                       &                       & S2 & 85.5 & 66.9 & 91.6 & 99.1 & 93.0 & 81.1 & 58.3 & 95.1 & 42.9 & \multirow{-2}{*}{30.2} \\
    \cmidrule{2-15}
    & \multirow{2}{*}{AutoScale}         & \multirow{2}{*}{85K}  & \multirow{2}{*}{50K}  & S1 & 97.4 & 85.1 & 99.6 & 100 & 83.5 & 96.7 & 96.4 & 97.8 & 76.4 &  \\
    &                                    &                       &                       & S2 & 84.4 & 69.3 & 92.9 & 98.6 & 88.3 & 79.8 & 58.2 & 96.4 & 45.9 & \multirow{-2}{*}{31.5} \\
    \cmidrule{2-15}
    & \multirow{2}{*}{AutoScale}         & \multirow{2}{*}{85K}  & \multirow{2}{*}{100K} & S1 & 98.1 & 87.8 & 98.9 & 99.3 & 83.8 & 97.1 & 96.2 & 97.6 & 80.9 &  \\
    &                                    &                       &                       & S2 & 88.0 & 73.4 & 93.1 & 99.1 & 86.4 & 82.7 & 56.6 & 94.9 & 46.9 & \multirow{-2}{*}{\textbf{33.3}} \\
    \midrule
    \multirow{10}{*}{DiffusionDrive~\cite{liao2025diffusiondrive}}
    & \multirow{2}{*}{Real Data}          & \multirow{2}{*}{85K}  & \multirow{2}{*}{-}    & S1 & 96.8 & 86.0 & 98.8 & 99.3 & 84.0 & 95.8 & 96.7 & 97.6 & 79.6 &  \\
    &                                    &                       &                       & S2 & 80.1 & 72.8 & 84.4 & 98.4 & 85.9 & 76.6 & 46.4 & 96.3 & 72.8 & \multirow{-2}{*}{27.5} \\
    \cmidrule{2-15}
    & \multirow{2}{*}{SimScale-recovery} & \multirow{2}{*}{85K}  & \multirow{2}{*}{147K} & S1 & 97.2 & 88.4 & 99.1 & 99.8 & 83.9 & 96.0 & 96.7 & 97.6 & 76.9 &  \\
    &                                    &                       &                       & S2 & 82.4 & 67.7 & 89.1 & 98.6 & 89.0 & 77.6 & 53.8 & 95.2 & 46.8 & \multirow{-2}{*}{30.4} \\
    \cmidrule{2-15}
    & \multirow{2}{*}{SimScale-planner}  & \multirow{2}{*}{85K}  & \multirow{2}{*}{237K} & S1 & 97.2 & 88.7 & 99.3 & 99.3 & 82.8 & 96.9 & 98.0 & 97.3 & 59.6 &  \\
    &                                    &                       &                       & S2 & 82.4 & 72.1 & 92.9 & 98.5 & 92.1 & 80.6 & 60.8 & 95.4 & 31.9 & \multirow{-2}{*}{32.6} \\
    \cmidrule{2-15}
    & \multirow{2}{*}{AutoScale}         & \multirow{2}{*}{85K}  & \multirow{2}{*}{50K}  & S1 & 97.2 & 86.2 & 99.2 & 99.3 & 84.4 & 96.4 & 97.3 & 97.6 & 80.0 &  \\
    &                                    &                       &                       & S2 & 87.8 & 71.2 & 91.5 & 98.6 & 89.7 & 82.2 & 61.1 & 95.6 & 42.1 & \multirow{-2}{*}{32.4} \\
    \cmidrule{2-15}
    & \multirow{2}{*}{AutoScale}         & \multirow{2}{*}{85K}  & \multirow{2}{*}{100K} & S1 & 97.9 & 89.1 & 99.2 & 99.3 & 84.3 & 97.1 & 98.0 & 97.8 & 77.8 &  \\
    &                                    &                       &                       & S2 & 83.4 & 72.7 & 91.6 & 98.8 & 90.0 & 77.4 & 61.2 & 94.6 & 46.1 & \multirow{-2}{*}{\textbf{33.5}} \\
    \bottomrule[1.2pt]
  \end{tabular}%
}
\end{table}

\begin{table}[t!]
  \centering
  \caption{Comparison of Real-Synthetic Co-Training in the NAVSIM-v2 \texttt{navtest} Benchmark.}
  \label{tab:real_synthetic_cotraining_navtest}
  \resizebox{\linewidth}{!}{%
  \begin{tabular}{l|c|c|c|cccc|ccccc|>{\columncolor{tablelightgray}}c}
    \toprule[1.2pt] 
    \textbf{Model} & \textbf{Settings} & \textbf{Real} & \textbf{Synthetic} 
    & \textbf{NC↑} & \textbf{DAC↑} & \textbf{DDC↑} & \textbf{TLC↑} 
    & \textbf{EP↑} & \textbf{TTC↑} & \textbf{LK↑} & \textbf{HC↑} & \textbf{EC↑} & \cellcolor{white}\textbf{EPDMS↑} \\
    \midrule
    \multirow{5}{*}{TransFuser~\cite{Chitta2023transfuser}}
    & Real Data          & 85K & -    & 97.7 & 94.0 & 99.3 & 99.8 & 87.2 & 96.7 & 95.5 & 98.3 & 82.9 & 81.5 \\
    & SimScale-recovery & 85K & 147K & 97.2 & 92.8 & 99.6 & 99.8 & 88.8 & 96.4 & 96.7 & 98.3 & 86.8 & 83.6 \\
    & SimScale-planner  & 85K & 237K & 98.3 & 95.6 & 99.6 & 99.8 & 87.1 & 97.5 & 97.2 & 98.3 & 88.2 & 84.4 \\
    & AutoScale         & 85K & 50K  & 98.1 & 95.8 & 99.4 & 99.9 & 86.7 & 97.5 & 97.4 & 98.3 & 87.3 & 84.1 \\
    & AutoScale         & 85K & 100K  & 97.8 & 96.1 & 99.7 & 99.8 & 88.3 & 97.0 & 97.0 & 98.3 & 87.6 & \textbf{84.5} \\
    \midrule
    \multirow{5}{*}{DiffusionDrive~\cite{liao2025diffusiondrive}}
    & Real Data          & 85K & -    & 98.4 & 95.5 & 99.5 & 99.8 & 87.5 & 97.5 & 96.9 & 98.4 & 87.7 & 84.2 \\
    & SimScale-recovery & 85K & 147K & 98.4 & 96.7 & 99.6 & 99.8 & 87.6 & 97.5 & 97.5 & 98.3 & 87.1 & 85.4 \\
    & SimScale-planner  & 85K & 237K & 98.5 & 97.1 & 99.6 & 99.8 & 87.4 & 97.8 & 98.0 & 98.3 & 87.5 & \textbf{85.9} \\
    & AutoScale         & 85K & 50K  & 98.5 & 96.8 & 99.6 & 99.8 & 87.4 & 97.7 & 97.6 & 98.3 & 87.7 & 85.6 \\
    & AutoScale         & 85K & 100K  & 98.5 & 97.0 & 99.6 & 99.8 & 87.5 & 97.7 & 97.4 & 98.3 & 87.9 & \textbf{85.9} \\
    \bottomrule[1.2pt] 
  \end{tabular}%
  }
\vspace{-3mm}
\end{table}

\subsection{Experimental Setup}
\label{sec:exp_setup}

\textbf{Real and Synthetic Data.}\quad
We use the \texttt{navtrain} training set from NAVSIM~\cite{Dauner2024navsim} as real data, comprising approximately 85K driving scenes.
We adopt SimScale~\cite{tian2025simscale}, a reactive simulation framework that reconstructs 3DGS scenarios, generates new trajectories and rerenders sensor data.
SimScale fully covers the \texttt{navtrain} training set, yielding a 384K synthetic pool.
We construct a 20K held-out calibration set via cluster-stratified sampling from the synthetic pool for closed-loop evaluation in simulation, and use the remainder as our synthetic candidate pool.

\textbf{Implementation Details.}\quad
In AutoScale, we set synthetic budget $B=50\text{K}$ or $100\text{K}$.
For Graph-RAE, we set $T_{\mathrm{len}}=12$ with $T_h=8$ and $T_f=4$.
We set regularization weight $\lambda=5$ and embedding dimension $128$, which is reduced to $64$ via PCA for GMM clustering.
For Cluster-GA, we set cluster number $K=64$, and run $T=5$ optimization rounds with half-cosine step schedule ($\epsilon_{\max}=0.1$).
In each round, we conduct closed-loop evaluation on the calibration set, and average results from the last 10 checkpoints to mitigate training randomness.
RBF bandwidth is set to $\sigma=0.5$, and ridge regularization coefficient to $\lambda_{\mathrm{reg}}=1.0$.

\textbf{Details of Baselines.}\quad
To illustrate the effectiveness of our AutoScale data engine, we compare it with vanilla co-training methods using larger synthetic data volumes and mainstream cross-domain data mixture methods on the real-synthetic co-training task.
For a fair comparison and to ensure performance differences are solely attributable to different data mixture strategies, all experiments adopt the same pre-generated synthetic data pool and follow unified model training protocols.

We compare with the following data mixture strategies:
(1) SimScale generates 147K recovery-based and 237K planner-based synthetic scenes, and directly co-trains them with 85K real driving data separately.
(2) Uniform: it selects data uniformly from the synthetic pool under a given budget.
(3) Chameleon~\cite{xie2025chameleon}: a data mixture framework for LLM pretraining and finetuning that computes domain weights via kernel ridge leverage scores in the latent space.
(4) IWR~\cite{xie2025iwr}: a retrieval method for robotic imitation learning using latent-space importance weights and feature distances.
We employ a shared Graph-RAE latent space for all cross-domain data mixture methods in autonomous driving.

All experiments are conducted on representative end-to-end planning models: the regression-based Transfuser (LTF)~\cite{Chitta2023transfuser} and the diffusion-based DiffusionDrive~\cite{liao2025diffusiondrive}.
We adopt their official implementations with two minor modifications: the input image resolution is unified to $2048 \times 512$, and LiDAR inputs are removed.
All models are trained from scratch on NVIDIA A800 GPUs.

\textbf{Benchmark and Metrics.}\quad
We adopt two official benchmarks from NAVSIMv2~\cite{Cao2025navsimv2} for evaluation: \texttt{navtest} and \texttt{navhard}.
\texttt{navtest} is a one-stage evaluation benchmark containing 12,146 diverse real-world driving scenarios.
\texttt{navhard} is an officially additionally collected two-stage challenging evaluation benchmark. It consists of 244 real-world scenarios in the first stage and 4,164 corresponding synthetic scenarios generated via 3DGS in the second stage.
The main evaluation metric is the Extended Predictive Driver Model Score (EPDMS).

\begin{table}[t!]
  \centering
  \scriptsize
  \caption{Data Mixture Method Comparison on the NAVSIM-v2 \texttt{navhard} Benchmark.}
  \label{tab:data_mixture_method_comparison_navhard}
  \resizebox{0.9\linewidth}{!}{%
  \begin{tabular}{l|c|cccc|ccccc|>{\columncolor{tablelightgray}}c}
    \toprule[1.2pt]
    \textbf{Method} & \textbf{Stage} & \textbf{NC↑} & \textbf{DAC↑} & \textbf{DDC↑} & \textbf{TLC↑} 
    & \textbf{EP↑} & \textbf{TTC↑} & \textbf{LK↑} & \textbf{HC↑} & \textbf{EC↑} & \cellcolor{white}\textbf{EPDMS↑} \\
    \midrule
    \multirow{2}{*}{Uniform}          & S1 & 97.6 & 86.7 & 99.1 & 99.6 & 84.3 & 96.0 & 97.3 & 97.6 & 78.7 &  \\
                                      & S2 & 83.2 & 70.3 & 91.3 & 98.3 & 89.9 & 77.3 & 57.7 & 95.4 & 44.6 & \multirow{-2}{*}{29.5} \\
    \cmidrule{1-12}
    \multirow{2}{*}{Chameleon~\cite{xie2025chameleon}} & S1 & 98.1 & 87.3 & 98.9 & 99.3 & 84.0 & 97.3 & 98.0 & 97.8 & 80.4 &  \\
                                      & S2 & 82.9 & 71.0 & 90.5 & 98.9 & 90.3 & 76.6 & 60.3 & 95.0 & 42.5 & \multirow{-2}{*}{30.9} \\
    \cmidrule{1-12}
    \multirow{2}{*}{IWR~\cite{xie2025iwr}}  & S1 & 98.2 & 86.9 & 99.6 & 99.8 & 83.6 & 95.6 & 96.7 & 97.6 & 77.8 &  \\
                                      & S2 & 84.3 & 73.3 & 91.0 & 98.6 & 88.6 & 79.1 & 57.1 & 96.2 & 44.6 & \multirow{-2}{*}{31.0} \\
    \cmidrule{1-12}
    \multirow{2}{*}{AutoScale}         & S1 & 97.2 & 86.2 & 99.2 & 99.3 & 84.4 & 96.4 & 97.3 & 97.6 & 80.0 &  \\
                                      & S2 & 87.8 & 71.2 & 91.5 & 98.6 & 89.7 & 82.2 & 61.1 & 95.6 & 42.1 & \multirow{-2}{*}{\textbf{32.4}} \\
    \bottomrule[1.2pt]
  \end{tabular}%
}
\vspace{-4mm}
\end{table}

\subsection{Main Results}

\textbf{Real-Synthetic Co-Training Results.}
We compare our method with the vanilla real-synthetic co-training strategy.
As shown in \cref{tab:real_synthetic_cotraining_navhard}, for LTF, our method achieves better EPDMS scores than SimScale-recovery and SimScale-planner, using only 34\% and 21\% additional synthetic data, respectively.
For DiffusionDrive, our method similarly outperforms the vanilla co-training strategy with a larger training budget by +3.1 and +0.9 in EPDMS.
In \cref{tab:real_synthetic_cotraining_navtest}, our AutoScale achieves comparable results to the vanilla real-synthetic co-training strategy on both LTF and DiffusionDrive with far less additional data.
Evaluations on two distinct models across \texttt{navhard} and \texttt{navtest} demonstrate that AutoScale not only effectively boosts model performance in challenging scenarios but also improves generalization ability, achieving higher data efficiency via our data mixture strategy.

\textbf{Data Mixture Method Comparison.}
We further compare our AutoScale with several cross-domain data mixture strategies on the DiffusionDrive model, under a fixed 50K synthetic data budget and identical training protocols.
As shown in \cref{tab:data_mixture_method_comparison_navhard}, when equipped with our Graph-RAE embedding for driving scene representation, both Chameleon and IWR outperform the uniform baseline under the same synthetic data budget.
This further verifies the effectiveness and generalization capability of our proposed Graph-RAE.
Our AutoScale surpasses all baseline data mixture approaches, demonstrating the superiority of our targeted, evaluation-driven closed-loop data mixture scheme.

\subsection{Ablation Studies and Discussions}

\textbf{Ablation Study on Graph-RAE.}
We ablate the regularizations for contrastive and metric learning in Graph-RAE separately and run the complete downstream closed-loop pipeline.
Experimental results under the shared DiffusionDrive training protocols and a 50K budget in \cref{tab:ablation_study} show that, compared with the full method, the model exhibits EPDMS performance drops of 0.2 and 0.3, respectively.
This demonstrates that representation quality affects clustering and retrieval performance.

\begin{figure}[t!]
    \centering
    \includegraphics[width=1.0\linewidth]{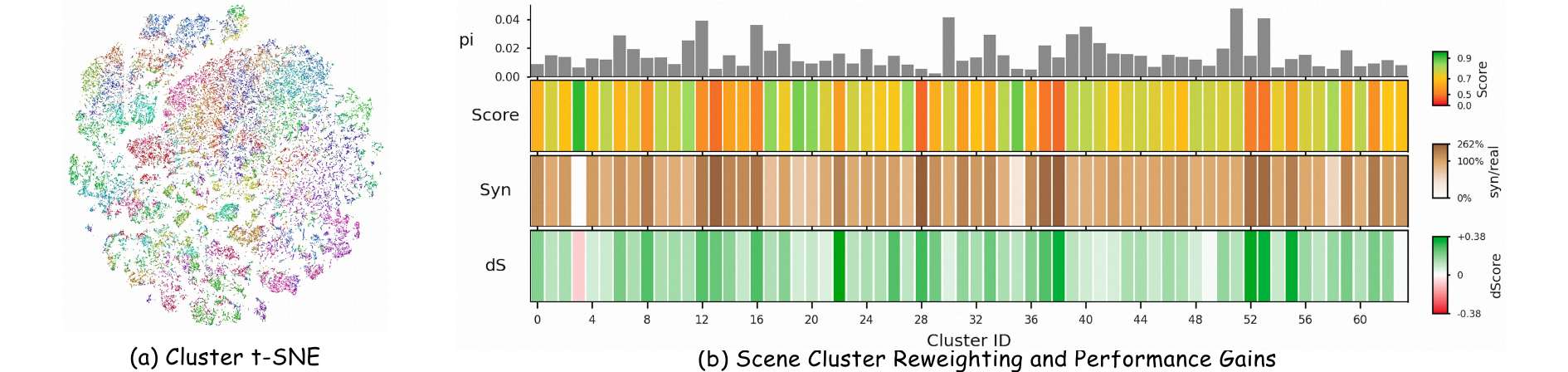}
    \vspace{-4mm}
    \caption{Qualitative results of clustering.
    (a) t-SNE of clusters on the training set.
    (b) Visualization of the scene cluster reweighting process and corresponding performance gains. 
    From top to bottom: cluster distribution of the training set, EPDMS scores of calibration set scenes assigned to clusters of the original model, synthetic-to-real data ratio, and per-cluster performance gains after data mixture.}
    \label{fig:cluster}
\end{figure}

\textbf{Effectiveness of Cluster-GA.}
We ablate the Cluster-GA module and perform pure retrieval with Graph-RAE representations. Experimental results in \cref{tab:ablation_study} show the efficacy of the Cluster-GA module.
We also investigate the selection of cluster numbers while keeping all other experimental settings unchanged. Setting the number of clusters to $K=64$ achieves the best model performance.
We also provide qualitative results.
In \cref{fig:cluster} (a), despite complex driving scenes and a large number of clusters, t-SNE visualizations present a compact and well-separated structure.
In \cref{fig:cluster} (b), low-performance clusters of the original model receive more synthetic data, and most of them achieve obvious performance gains after data mixture.

\begin{figure}[t!]
    \centering
    \includegraphics[width=1.0\linewidth]{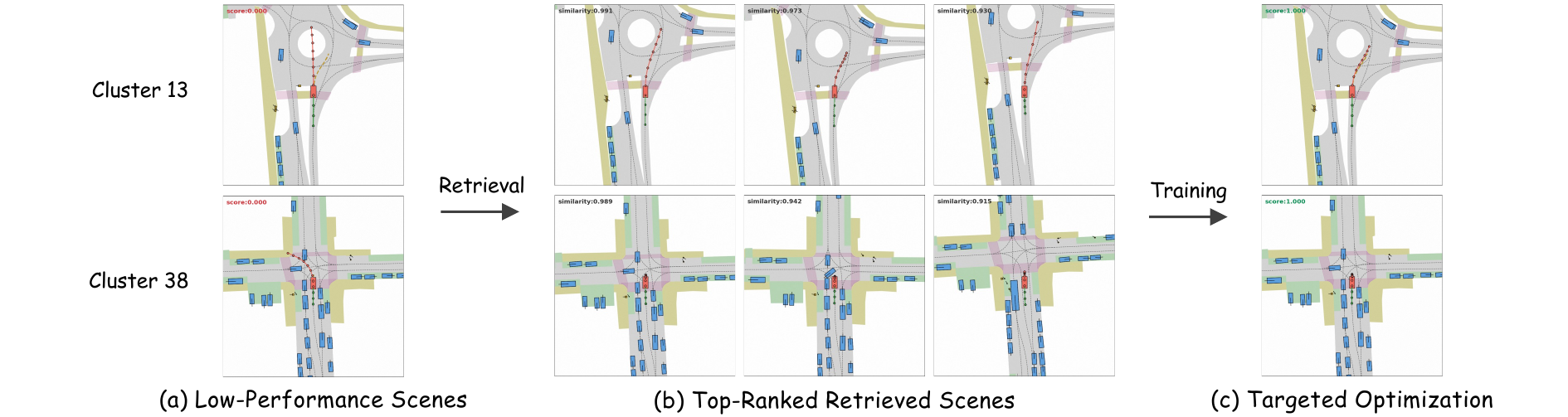}
    \vspace{-4mm}
    \caption{Qualitative results of retrieval.
    The green dotted curve, red dotted curve, and dark yellow dashed line represent historical and future trajectories, and ground truth, respectively.}
    \label{fig:retrieval}
    \vspace{-4mm}
\end{figure}

\textbf{Effectiveness of Retrieval.}
We ablate the retrieval module and perform vector-based clustering and reweighting. In each round, we assign synthetic data to clusters and uniformly sample data per cluster. Experimental results in \cref{tab:ablation_study} show the efficacy of the retrieval module.
We also provide qualitative results.
In \cref{fig:retrieval}, we select representative low-performance scenes from the original model corresponding to \cref{fig:cluster} (b), where the ego vehicle performs poorly when navigating roundabouts and makes improper left turns at intersections, yielding an EPDMS score of 0 (shown in the top-left corner).
We effectively retrieve the most similar synthetic data. After data mixture and retraining, the performance of these two scenes is significantly improved.

\begin{table}[t!]
  \centering
  \scriptsize 
  \caption{Ablation Study Results on the NAVSIM-v2 \texttt{navtest} Benchmark.}
  \label{tab:ablation_study}
  \resizebox{0.9\linewidth}{!}{
  \begin{tabular}{l|c|cccc|ccccc|>{\columncolor{tablelightgray}}c}
    \toprule[1.2pt]
    \textbf{Ablation} & \textbf{Setting} & \textbf{NC↑} & \textbf{DAC↑} & \textbf{DDC↑} & \textbf{TLC↑}
    & \textbf{EP↑} & \textbf{TTC↑} & \textbf{LK↑} & \textbf{HC↑} & \textbf{EC↑} & \cellcolor{white}\textbf{EPDMS↑} \\
    \midrule
    \multirow{2}{*}{Abl.module}
    & w/o Cluster-GA & 98.3 & 96.6 & 99.6 & 99.8 & 87.5 & 97.7 & 97.5 & 98.3 & 87.2 & 85.2 \\
    & w/o Retrieval  & 98.5 & 96.6 & 99.6 & 99.8 & 87.5 & 97.6 & 97.5 & 98.3 & 87.4 & 85.4 \\
    \midrule
    \multirow{2}{*}{Abl.vector}
    & w/o Contrastive Reg. & 98.3 & 96.7 & 99.6 & 99.8 & 87.5 & 97.6 & 97.7 & 98.3 & 87.6 & 85.4 \\
    & w/o Metric Reg. & 98.3 & 96.6 & 99.6 & 99.8 & 87.6 & 97.5 & 97.6 & 98.3 & 87.6 & 85.3 \\
    \midrule
    \multirow{2}{*}{Abl.cluster}
    & 32  & 98.5 & 96.7 & 99.7 & 99.8 & 87.5 & 97.6 & 97.4 & 98.3 & 87.5 & 85.4 \\
    & 128 & 98.3 & 96.7 & 99.6 & 99.8 & 87.5 & 97.6 & 97.7 & 98.4 & 87.6 & 85.4 \\
    \midrule
    AutoScale & Full & 98.5 & 96.8 & 99.6 & 99.8 & 87.4 & 97.7 & 97.6 & 98.3 & 87.7 & \textbf{85.6} \\
    \bottomrule[1.2pt]
  \end{tabular}
  }
\end{table}

\section{Conclusion and Limitation}
\vspace{-3mm}
In this paper, we pioneer a systematic investigation into the data mixture effect on model performance in real-synthetic co-training.
We argue that training data mixture requires explicit guidance on scene types and quantities, and formulate it as an iterative optimization process to improve model evaluation performance.
We propose AutoScale, a closed-loop dynamic data engine unifying scene representation, data mixture optimization and retrieval, and model training and evaluation.
Experiments on NavSim show that, guided by evaluation in simulation, AutoScale achieves better real-world performance with fewer synthetic samples.
To keep our investigation focused, we only conduct experiments on two representative models with limited iterative rounds.
Future work includes extending the data mixture effect to more model architectures (e.g., world action models) and training objectives (e.g., reinforcement learning), and accumulating more training-evaluation rounds naturally in industrial development workflows to precisely estimate its impact on model performance.


\clearpage
\bibliographystyle{plain}
\bibliography{main}


\clearpage
\appendix
\section*{Technical appendices and supplementary material}

This supplementary material presents implementation and evaluation details of our method and baselines in \cref{sec:implementation_details}.
We further provide experimental results and analyses in \cref{sec:additional_exp}.
Supplementary qualitative visualization results are presented in \cref{sec:additional_viz}.

\section{Additional Implementation and Evaluation Details}
\label{sec:implementation_details}

\subsection{Additional Implementation Details of Graph-RAE}

\textbf{Graph Definition.}
Each driving scenario is constructed as a heterogeneous graph $\mathcal{G}=(\mathcal{V},\mathcal{E})$,
where $\mathcal{V}$ and $\mathcal{E}$ denote node set and edge set, respectively.
All graph nodes are divided into two categories: dynamic nodes $\mathcal{V}_{\text{dyn}}$ including the ego-vehicle and surrounding traffic participants, and static map nodes $\mathcal{V}_{\text{map}}$ including pedestrian crossings, road dividers and boundary elements.
Each node $v_i$ contains a $T_{\mathrm{len}}$-step sequential polyline feature $\mathbf{V}_i$. The per-time-step node feature is defined as:
\begin{equation}
\mathbf{v}_{i,t} =
\begin{cases}
[x,y,\theta,\kappa,\nu]^\top\!\in\mathbb{R}^5, & v_i\in\mathcal{V}_{\text{dyn}},\\
[x,y,\theta,\kappa]^\top\!\in\mathbb{R}^4, & v_i\in\mathcal{V}_{\text{map}},
\end{cases}
\end{equation}
where $x,y$ denote 2D position, $\theta$ denotes heading angle, $\kappa$ denotes trajectory curvature, and $\nu$ denotes velocity only for dynamic nodes.
$\mathbf{E}_{ji}$ represents the $T_{\mathrm{len}}$-step relative spatio-temporal feature sequence of each directed edge.
$\mathbf{emb}_{\text{pos}}$ denotes temporal positional embedding, and $\mathbf{emb}_{\text{sem}}$ denotes unified semantic embedding for node and edge category attributes.

\textbf{Feature Encoding.}
All raw node and edge sequential features are projected by dedicated MLPs, augmented with only positional and semantic embeddings to obtain unified hidden representations:
\begin{equation}
\mathbf{h}_i = \text{MLP}(\mathbf{V}_i) + \mathbf{emb}_{\text{sem}} + \mathbf{emb}_{\text{pos}},
\quad
\mathbf{e}_{ji} = \text{MLP}(\mathbf{E}_{ji}) + \mathbf{emb}_{\text{pos}}.
\end{equation}

\textbf{Graph Transformer Propagation.}
We stack $L$ layers of graph transformer with gated multi-head attention to aggregate spatio-temporal interaction information from neighboring nodes and connected edges. The simplified layer-wise node representation update rule is formulated as:
\begin{equation}
\mathbf{h}_i^{(l+1)} = \mathbf{h}_i^{(l)} + \sigma\!\big(\mathbf{g}_i^{(l)}\big) \odot \mathrm{MHA}\!\big(\mathbf{h}_i^{(l)},\{\mathbf{h}_j^{(l)},\mathbf{e}_{ji}\}\big).
\end{equation}
where $\mathbf{g}_i^{(l)}$ is the learnable gate vector at layer $l$, and $\sigma(\cdot)$ denotes the sigmoid activation function.
After multi-layer feature refinement, the ego-vehicle node embedding is $\ell_2$-normalized to generate the final global scene representation:
\begin{equation}
\mathbf{z}_{\text{scene}} = \mathbf{h}_{\text{ego}}^{(L)} / \|\mathbf{h}_{\text{ego}}^{(L)}\|_2.
\end{equation}

\textbf{Reconstruction Loss.}
The graph decoder reconstructs the original polyline sequences for all nodes, and the autoencoder training is optimized with standard Huber loss:
\begin{equation}
\mathcal{L}_{\text{recon}} = \sum_{v_i}\sum_{t=1}^{T_{\mathrm{len}}}\text{Huber}(\hat{\mathbf{v}}_{i,t},\mathbf{v}_{i,t}).
\end{equation}

\textbf{Contrastive Regularization.}
Two contrastive learning strategies are adopted for embedding space regularization, including SimCLR~\cite{chen2020simclr} for geometric invariance and SupCon~\cite{khosla2020supcon} for semantic clustering. The total contrastive objective is 
$\mathcal{L}_{\text{contra}}=\mathcal{L}_{\text{simclr}}+\mathcal{L}_{\text{supcon}}$:
\begin{equation}
\mathcal{L}_{\text{simclr}} = -\frac{1}{N}\sum\log\frac{\exp(\text{sim}(\mathbf{z}^a,\mathbf{z}^b)/\tau)}{\exp(\text{sim}(\mathbf{z}^a,\mathbf{z}^b)/\tau)+\sum_{m\neq k}\exp(\text{sim}(\mathbf{z}^a,\mathbf{z}^m)/\tau)},
\end{equation}
\begin{equation}
\mathcal{L}_{\text{supcon}} = -\frac{1}{N}\sum\frac{1}{|P_k|}\sum\log\frac{\exp(\text{sim}(\mathbf{z},\mathbf{z}_p)/\tau)}{\sum_{m\neq k}\exp(\text{sim}(\mathbf{z},\mathbf{z}_m)/\tau)}.
\end{equation}
where $N$ is the batch size, $\tau$ denotes the temperature coefficient, $a,b$ denote two geometrically augmented views of the same scene, $k,m$ index individual scenes, $P_k$ denotes the positive sample set sharing consistent semantic labels with scene $k$, and $p$ enumerates positive samples within $P_k$.

\textbf{Metric Learning Regularization.}
A pairwise metric regression objective is further added to align scene embeddings with actual driving task performance:
\begin{equation}
\mathcal{L}_{\text{metric}} = \frac{1}{N^2}\sum_{i,j}\mathbb{I}(i,j)\cdot\big\|\hat{|\mathbf{m}_i-\mathbf{m}_j|}-|\mathbf{m}_i-\mathbf{m}_j|\big\|_2^2.
\end{equation}
where $\mathbb{I}(i,j)$ is the valid pair indicator function, $\mathbf{m}_i$ denotes the ground-truth evaluation metric vector, and $\hat{\cdot}$ represents the predicted counterpart.

\textbf{Overall Training Loss.}
The final joint training loss balances reconstruction fitting and two regularization tasks with a weight coefficient $\lambda$:
\begin{equation}
\mathcal{L} = \mathcal{L}_{\text{recon}} + \lambda\big(\mathcal{L}_{\text{contra}} + \mathcal{L}_{\text{metric}}\big).
\end{equation}

\subsection{Implementation Details of Baselines}

\textbf{IWR~\cite{xie2025iwr}.}
We use the same implementation details as the original paper.
The core idea of IWR is to retain synthetic samples whose embedding distribution most closel resembles that of the real training data.
Given real embeddings $\{\mathbf{v}_x\}_{x\in\mathcal{D}_{\mathrm{real}}}$ and synthetic embeddings $\{\mathbf{v}_x\}_{x\in\mathcal{D}_{\mathrm{syn}}}$, we first perform a joint PCA on their concatenation and project both sets to $n{=}16$ dimensions, alleviating the curse of dimensionality for subsequent density estimation.
A Gaussian KDE is then fitted separately on the projected real and synthetic embeddings, using Scott's bandwidth $h=N_0^{-1/(n+4)}$.
For every synthetic sample $x\in\mathcal{D}_{\mathrm{syn}}$, the importance weight $w_x = \exp\!\bigl(\log \hat{p}_{\mathrm{real}}(\mathbf{v}_x) - \log \hat{p}_{\mathrm{syn}}(\mathbf{v}_x)\bigr)$ measures how much more likely the sample is under the real distribution than under the synthetic one; log-weights are clipped to $[-100,100]$ for numerical stability.
The final subset $\mathcal{D}^{*}$ is formed by selecting the $B$ synthetic samples with the highest importance weights, effectively prioritising synthetic scenes that fill gaps in the real data distribution.

\textbf{Chameleon~\cite{xie2025chameleon}.}
We use the same implementation details as the original paper.
Chameleon partitions the synthetic pool into $K$ clusters defined by cluster centroids, and allocates the budget $B$ unevenly across clusters based on how well each cluster is already covered by the others.
Concretely, we form the cluster similarity matrix $\mathbf{R} = \tilde{\mathbf{C}}\tilde{\mathbf{C}}^{\top}\in\mathbb{R}^{K\times K}$, where $\tilde{\mathbf{C}}$ is the $\ell_2$-normalized centroid matrix and $\mathbf{R}_{kj}$ reflects the
cosine similarity between cluster $k$ and cluster $j$.
KRLS leverage scores $\gamma_k = \bigl[\mathbf{R}(\mathbf{R} + K\lambda \mathbf{I})^{-1}\bigr]_{kk}$ are then derived from this matrix: a low score signals that cluster $k$ is underrepresented and cannot be well explained by the remaining clusters.
Cluster weights $\bar{w}_k \propto \exp(1/\gamma_k)$ therefore assign a larger synthetic allocation to underrepresented clusters, and the per-cluster budget is set to $c_k = \operatorname{round}(\bar{w}_k B)$ with a greedy integer-correction step that enforces $\sum_k c_k = B$ exactly.
Finally, each synthetic sample $x\in\mathcal{D}_{\mathrm{syn}}$ is assigned to its nearest centroid $\mathbf{c}_k$, $c_k$ samples are drawn uniformly at random from each cluster pool $\mathcal{P}_k$, and any shortfall caused by insufficient pool size is filled by uniform sampling from $\mathcal{D}_{\mathrm{syn}}\setminus\mathcal{D}^{*}$.

\subsection{Model and Training}

\textbf{LTF~\cite{Chitta2023transfuser}.} 
A regression-based planner that directly regresses future waypoints from fused sensory latents. 
We adopt the same implementation details as those in SimScale~\cite{tian2025simscale},
it employs a pre-trained ResNet34~\cite{he2016resnet} as the image encoder for 3 camera inputs (front, front-left, front-right). The input images are concatenated to a total resolution of $2048 \times 512$ for feature extraction. 
The model is trained for 100 epochs on 8 GPUs with a total batch size of 512, using the Adam~\cite{kingma2015adam} optimizer with a learning rate of $2.83 \times 10^{-4}$.

\textbf{DiffusionDrive~\cite{liao2025diffusiondrive}.} 
A multi-modal generative planner that iteratively denoises diverse trajectories via anchor-conditioned truncated diffusion. 
Each anchor adaptively queries the encoded image features. 
We follow the identical experimental configurations outlined in SimScale~\cite{tian2025simscale},
it employs a pre-trained ResNet34~\cite{he2016resnet} as the image encoder for 3 camera inputs (front, front-left, front-right). The input images are concatenated to a total resolution of $2048 \times 512$ for feature extraction, and truncated diffusion with 20 clustered anchors is applied. 
The model is trained for 100 epochs on 8 NVIDIA A800 GPUs with a total batch size of 512, using the AdamW~\cite{loshchilov2019adamw} optimizer with a learning rate of $6 \times 10^{-4}$.

\subsection{Additional Evaluation Metric Details}

We use both the Predictive Driver Model Score (PDMS) and the Extended Predictive Driver Model Score (EPDMS) for evaluation.

PDMS includes No At-Fault Collision (NC), Drivable Area Compliance (DAC), Time-To-Collision (TTC), Comfort (Comf.), and Ego Progress (EP). 
PDMS is defined as:
\begin{equation}
\text{PDMS} = \text{NC} \times \text{DAC} \times \frac{5\times\text{EP} + 5\times\text{TTC} + 2\times\text{Comf.}}{12}
\end{equation}

EPDMS includes No At-Fault Collision (NC), Drivable Area Compliance (DAC), Driving Direction Compliance (DDC), Traffic Light Compliance (TLC), Ego Progress (EP), Time to Collision (TTC), Lane Keeping (LK), History Comfort (HC), and Extended Comfort (EC). The four multiplicative terms (NC, DAC, DDC, TLC) act as strict safety constraints, while the weighted sum in the numerator comprehensively evaluates overall driving performance.
EPDMS is defined as:
\begin{equation}
  \text{EPDMS}
  = \text{NC} \times \text{DAC} \times \text{DDC}
    \times \text{TLC} \times
    \frac{
      5 \!\cdot\! \text{EP}
      + 5 \!\cdot\! \text{TTC}
      + 2 \!\cdot\! \text{LK}
      + 2 \!\cdot\! \text{HC}
      + 2 \!\cdot\! \text{EC}
    }{16}.
  \label{eq:epdms}
\end{equation}

\section{Additional Experimental Results and Discussion}
\label{sec:additional_exp}
\subsection{Additional Results for Real-Synthetic Co-Training}

\begin{table}[htbp]
  \centering
  \caption{Comparison of Real-Synthetic Co-Training on the NAVSIM-v1 \texttt{navtest} Benchmark.}
  \label{tab:real_synthetic_cotraining_navtestv1}
  \resizebox{\linewidth}{!}{%
  \begin{tabular}{l|c|c|c|ccccc|>{\columncolor{tablelightgray}}c}
    \toprule[1.2pt]
    \textbf{Model} & \textbf{Settings} & \textbf{Real} & \textbf{Synthetic} 
    & \textbf{NC↑} & \textbf{DAC↑} & \textbf{EP↑} & \textbf{TTC↑} & \textbf{Comf.↑} & \cellcolor{white}\textbf{PDMS↑} \\
    \midrule
    \multirow{5}{*}{TransFuser~\cite{Chitta2023transfuser}}
    & Real Data          & 85K & -    & 97.5 & 93.7 & 80.0 & 92.6 & 100 & 84.8 \\
    & SimScale-recovery & 85K & 147K & 98.2 & 95.0 & 80.5 & 94.3 & 100 & 86.4 \\
    & SimScale-planner  & 85K & 237K & 98.3 & 95.6 & 91.3 & 94.6 & 100 & 87.3 \\
    & AutoScale         & 85K & 50K  & 98.0 & 95.8 & 82.0 & 93.6 & 100 & 87.2 \\
    & AutoScale         & 85K & 100K & 98.0 & 96.1 & 82.2 & 93.6 & 100 & \textbf{87.5} \\
    \midrule
    \multirow{5}{*}{DiffusionDrive~\cite{liao2025diffusiondrive}}
    & Real Data          & 85K & -    & 98.2 & 95.2 & 81.5 & 94.4 & 100 & 87.1 \\
    & SimScale-recovery & 85K & 147K & 98.4 & 96.5 & 82.8 & 94.7 & 100 & 88.4 \\
    & SimScale-planner  & 85K & 237K & 98.5 & 97.1 & 83.1 & 94.7 & 100 & 88.9 \\
    & AutoScale         & 85K & 50K  & 98.5 & 96.7 & 82.9 & 94.8 & 100 & 88.7 \\
    & AutoScale         & 85K & 100K & 98.5 & 96.9 & 83.1 & 95.0 & 100 & \textbf{89.0} \\
    \bottomrule[1.2pt]
  \end{tabular}%
  }
\end{table}

We provide additional comparisons of our method with the vanilla real-synthetic co-training strategy NavSim-v1 \texttt{navtest}~\cite{Dauner2024navsim}.
In \cref{tab:real_synthetic_cotraining_navtestv1}, similar to the main text, our AutoScale achieves comparable results to the vanilla real-synthetic co-training strategy on both LTF and DiffusionDrive with far less additional data, demonstrating that AutoScale effectively boosts model performance and improves generalization ability, achieving higher data efficiency via our data mixture strategy.

\subsection{Additional Results for Data Mixture Methods Comparison}

\begin{table}[htbp]
  \centering
  \scriptsize
  \caption{Data Mixture Method Comparison in the NAVSIM-v2 \texttt{navtest} Benchmark.}
  \label{tab:data_mixture_method_comparison_navtestv2}
  \begin{tabular}{l|cccc|ccccc|>{\columncolor{tablelightgray}}c}
    \toprule[1.2pt]
    \textbf{Method} & \textbf{NC↑} & \textbf{DAC↑} & \textbf{DDC↑} & \textbf{TLC↑}
    & \textbf{EP↑} & \textbf{TTC↑} & \textbf{LK↑} & \textbf{HC↑} & \textbf{EC↑} & \cellcolor{white}\textbf{EPDMS↑} \\
    \midrule
    Uniform        & 98.4 & 96.2 & 99.5 & 99.8 & 87.6 & 97.6 & 97.5 & 98.3 & 87.5 & 84.9 \\
    Chameleon ~\cite{xie2025chameleon} & 98.3 & 96.5 & 99.5 & 99.8 & 87.5 & 97.5 & 97.5 & 98.3 & 87.4 & 85.2 \\
    IWR ~\cite{xie2025iwr} & 98.5 & 96.5 & 99.6 & 99.8 & 87.5 & 97.6 & 97.5 & 98.3 & 87.6 & 85.3 \\
    AutoScale      & 98.5 & 96.8 & 99.6 & 99.8 & 87.4 & 97.7 & 97.6 & 98.3 & 87.7 & \textbf{85.6} \\
    \bottomrule[1.2pt]
  \end{tabular}%
\end{table}

We provide additional comparisons of our AutoScale against several cross-domain data mixture strategies on the DiffusionDrive model. 
Evaluations are conducted on the NavSim-v2 \texttt{navtest} split~\cite{Dauner2024navsim}, with a fixed 50K synthetic data budget and identical training protocols.
As shown in \cref{tab:data_mixture_method_comparison_navtestv2}, consistent with the main text results, the experiments validate the effectiveness of our proposed Graph-RAE for driving scene representation, as well as the superiority of our data mixture scheme.

\subsection{Model-Specific Data Mixture}
\label{sec:model_specific}

We investigate whether the optimal data mixture in real-synthetic co-training is transferable across model architectures by quantifying mixture divergence and conducting cross-model experiments.

\textbf{Mixture Divergence.}
Given two token sets $\mathcal{A}$ and $\mathcal{B}$ selected for different models under an identical budget from the same synthetic pool, we measure their divergence via the Jaccard similarity~\cite{jaccard1901}:
\begin{equation}
J(\mathcal{A}, \mathcal{B}) = \frac{|\mathcal{A} \cap \mathcal{B}|}{|\mathcal{A} \cup \mathcal{B}|}.
\end{equation}
Under a budget of 100K, DiffusionDrive and Transfuser share approximately 57K tokens out of a combined 143K unique selections in the total 364K synthetic data pool, yielding $J = 0.399$. 
This indicates that over half of each model’s preferred synthetic data is unique to its own architecture.

\textbf{Cross-Model Training.}
To verify the data mixture divergence reflects performance impact, we conduct a swap experiment: each model is trained with the data mixture optimized for the other, with a budget of 100K and all hyperparameters kept identical.

\begin{table}[htbp]
  \centering
  \scriptsize
  \caption{Cross-Model Training Experiment on the NAVSIM-v2 \texttt{navtest} Benchmark.}
  \label{tab:cross_model}
  \begin{tabular}{l|c>{\columncolor{tablelightgray}}c}
    \toprule[1.2pt]
    \textbf{Model} & \textbf{Data Mixture} & \cellcolor{white}\textbf{EPDMS↑} \\
    \midrule
    \multirow{2}{*}{Transfuser}     & Cross-Optimized  & 84.4  \\
                                    & Self-Optimized & \textbf{84.5}  \\
    \midrule
    \multirow{2}{*}{DiffusionDrive} & Cross-Optimized  & 85.6  \\
                                    & Self-Optimized & \textbf{85.9}  \\
    \bottomrule[1.2pt]
  \end{tabular}%
\end{table}

As shown in Table~\ref{tab:cross_model}, each model is trained with either its own optimized data mixture (self) or that of the counterpart (cross).
All models yield the best performance under self-optimized mixtures, while cross-model settings lead to performance degradation. 
This verifies that optimal data mixtures are model-dependent, highlighting the necessity of tailored strategies for different model architectures.

\subsection{Effectiveness of Cross-Cluster Propagation}
\label{sec:cross_cluster}

We model the effect of data mixture changes on per-cluster performance using a relative fitting formulation in ~\eqref{eq:surrogate}.
The matrix $\mathbf{R}$ encodes inter-cluster similarity via an RBF kernel:
\begin{equation}
\mathbf{R}_{kj} = \exp\left(-\frac{d_{\cos}(k,j)^2}{2\sigma^2}\right),
\end{equation}
where $d_{\cos}(k,j) = 1 - \cos(\mathbf{c}_k, \mathbf{c}_j)$ is the cosine distance between cluster centroids.

To quantify whether cross-cluster propagation provides additional modeling power beyond independent per-cluster effects, we compare the coefficient of determination under two settings:
\begin{equation}
R^2 = 1 - \frac{\sum_{i}(\Delta s_i - \Delta\hat{s}_i)^2}{\sum_{i}(\Delta s_i - \overline{\Delta s})^2},
\end{equation}
evaluated with the full RBF kernel $\mathbf{R}$ versus the identity matrix $\mathbf{I}$ (no propagation, each cluster modeled independently).

\begin{table}[htbp]
  \centering
  \scriptsize
  \caption{Cross-Cluster Propagation Effectiveness on NAVSIM-v2 \texttt{navtest}. $R^2(\mathbf{R})$: full RBF kernel; $R^2(\mathbf{I})$: no propagation (identity matrix); $\Delta R^2$: performance improvement brought by propagation.}
  \label{tab:r_matrix}
  \begin{tabular}{l|cccc}
    \toprule[1.2pt]
    \textbf{Model} & \textbf{$R^2(\mathbf{R})$} & \textbf{$R^2(\mathbf{I})$} & \textbf{$\Delta R^2$} \\
    \midrule
    \multirow{1}{*}{DiffusionDrive} & 0.415 & 0.409 & +0.006 \\
    \multirow{1}{*}{Transfuser}     & 0.134 & 0.068 & +0.066 \\
    \bottomrule[1.2pt]
  \end{tabular}%
\end{table}

As shown in \cref{tab:r_matrix}, under a 100K budget, cross-cluster propagation improves both models but with distinct effects. 
It brings a marginal gain ($\Delta R^2 = +0.006$) for DiffusionDrive, while yielding a substantial boost ($\Delta R^2 = +0.066$) for Transfuser.

\clearpage

\section{Additional Qualitative Results}
\label{sec:additional_viz}

\begin{figure}[htbp]
    \centering
    \includegraphics[width=0.8\linewidth]{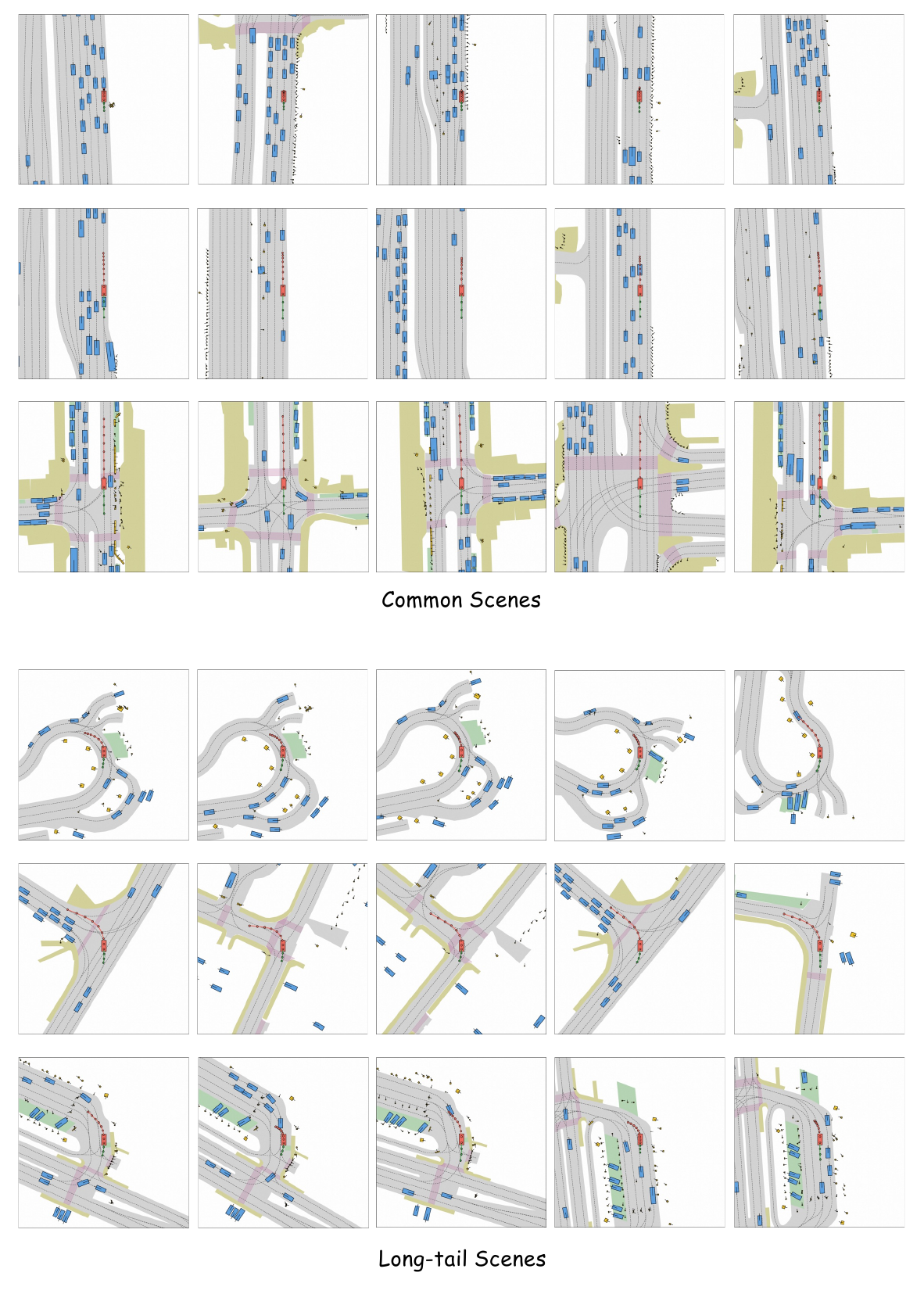}
    \caption{Clustering Visualization of Common and Long-Tail Driving Scenes in the \texttt{navtrain} Training Split of the NavSim Dataset.}
    \label{fig:appendix_cluster}
\end{figure}

\begin{figure}[htbp]
    \centering
    \includegraphics[width=1.0\linewidth]{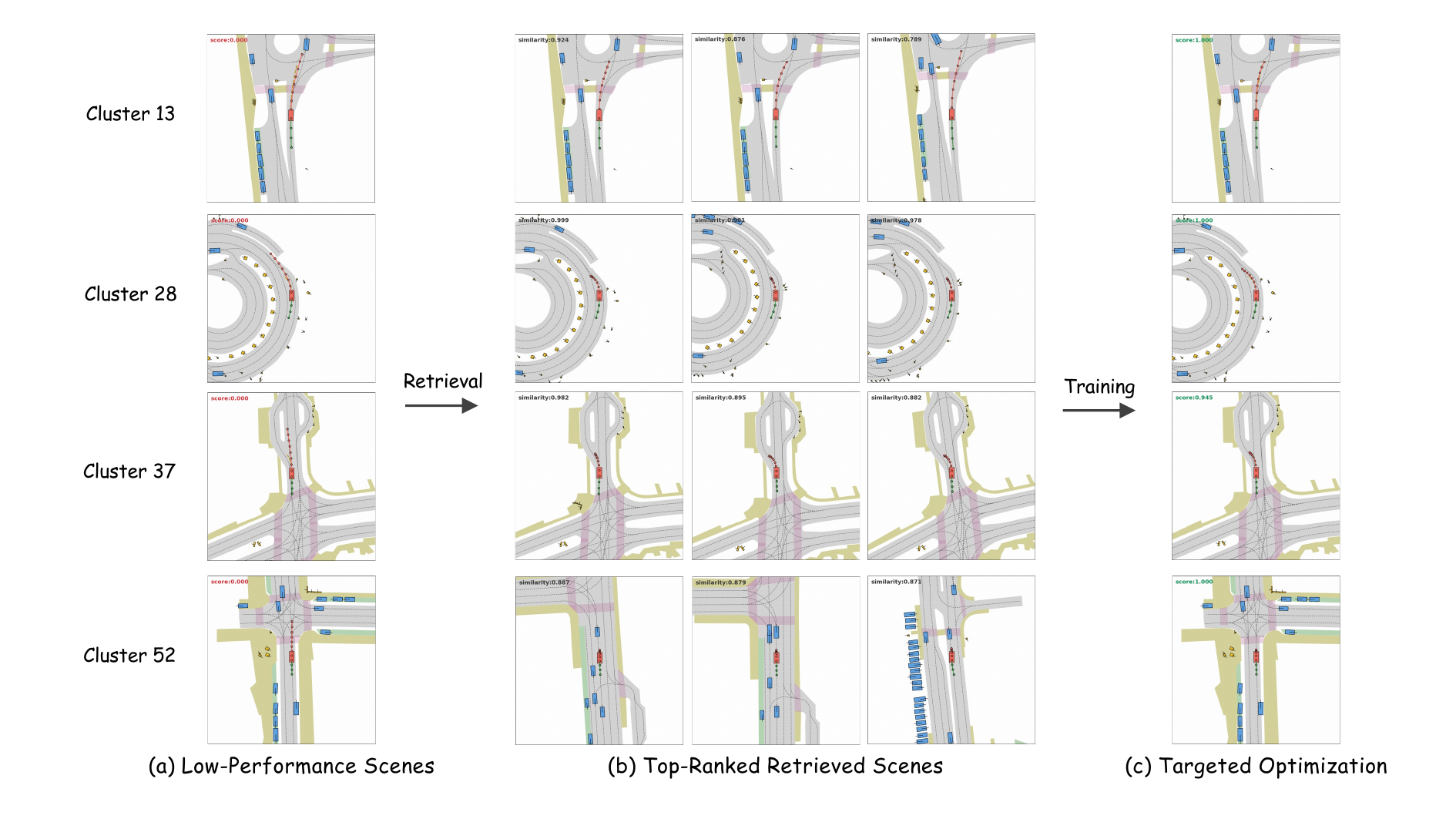}
    \caption{Additional Visualization of Underperforming Scenes, Scene Retrieval, and Performance Improvement.}
    \label{fig:appendix_retrieval}
\end{figure}

Corresponding to the clustering and related performance statistics in \cref{fig:cluster} (b), we further provide representative clustering results of common and long-tail scenes in the \texttt{navtrain} training split of NavSim in \cref{fig:appendix_cluster}, as well as additional visualizations of underperforming scenes, scene retrieval, and performance improvement in \cref{fig:appendix_retrieval}.

\clearpage

\clearpage


\end{document}